\documentclass{article}

\usepackage[preprint]{gpsmdms_2022}
\usepackage[utf8]{inputenc} % allow utf-8 input
\usepackage[T1]{fontenc}    % use 8-bit T1 fonts
\usepackage{hyperref}       % hyperlinks
\usepackage{url}            % simple URL typesetting
\usepackage{booktabs}       % professional-quality tables
\usepackage{amsfonts}       % blackboard math symbols
\usepackage{nicefrac}       % compact symbols for 1/2, etc.
\usepackage{microtype}      % microtypography
\usepackage{xcolor}         % colors
\usepackage{bm}
\usepackage{graphicx}
\usepackage{amsmath}
\usepackage{algorithm}
\usepackage[noend]{algpseudocode}

\DeclareMathOperator*{\argmin}{arg\,min}

\newcommand*{\figuretitle}[1]{%
    {\centering%   <--------  will only affect the title because of the grouping (by the
    \textbf{#1}%              braces before \centering and behind \medskip). If you remove
    \par\medskip}%            these braces the whole body of a {figure} env will be centered.
}

\title{Scalable Gaussian Process Hyperparameter Optimization via Coverage Regularization
}

\author{%
  Killian Wood  \\
  Department of Applied Mathematics\\
  University of Colorado Boulder\\
  Boulder, CO 80301 \\
  \texttt{killian.wood@colorado.edu} \\
  \And
  Alec M.~Dunton \\
  Center for Applied Scientific Computing\\
  Lawrence Livermore National Laboratory\\
  Livermore, CA 94550 \\
  \texttt{dunton1@llnl.gov} \\
  \And
  Amanda Muyskens \\
  Applied Statistics Group \\
  Lawrence Livermore National Laboratory\\
  Livermore, CA 94550 \\
  \texttt{muyskens1@llnl.gov} \\
  \And
  Benjamin W. Priest \\
  Center for Applied Scientific Computing\\
  Lawrence Livermore National Laboratory\\
  Livermore, CA 94550 \\
  \texttt{priest2@llnl.gov} \\
}

\begin{document}

\maketitle

\begin{abstract}
	Gaussian processes (GPs) are Bayesian non-parametric models popular in a variety of applications due to their accuracy and native uncertainty quantification (UQ).
	Tuning GP hyperparameters is critical to ensure the validity of prediction accuracy and uncertainty; uniquely estimating multiple hyperparameters in, e.g. the Mat\'ern kernel can also be a significant challenge. 
	Moreover, training GPs on large-scale datasets is a highly active area of research: traditional maximum likelihood hyperparameter training requires quadratic memory to form the covariance matrix and has cubic training complexity.
	To address the scalable hyperparameter tuning problem, we present a novel algorithm which estimates the smoothness and length-scale parameters in the Mat{\`e}rn kernel in order to improve robustness of the resulting prediction uncertainties. 
	Using novel loss functions similar to those in conformal prediction algorithms in the computational framework provided by the hyperparameter estimation algorithm MuyGPs, we achieve improved UQ over leave-one-out likelihood maximization while maintaining a high degree of scalability as demonstrated in numerical experiments.
\end{abstract}

\textbf{Introduction.} Gaussian process regression (GPR) approximates a function $f : \mathbb{R}^{d} \rightarrow \mathbb{R}^{\ell}$ using training points in $\mathbb{R}^{d} \times \mathbb{R}^{\ell}$.
We can think of these points as forming the rows of a matrix $\bm{X} \in \mathbb{R}^{n \times d}$, with observations $Y(\bm{X}) \in \mathbb{R}^{n \times \ell}$.
We assume that the target function $f$ is drawn from the  distribution $\mathcal{GP}(m(\bm{x}),k_\theta(\bm{x},\bm{x'}))$,
where $m(\bm{x})$ is the mean of the GP evaluated at the location $\bm{x}$.
$k_\theta(\bm{x},\bm{x'})$ is the {\it kernel} function, which generates the covariance between $\bm{x}$ and $\bm{x'}$ and is controlled by hyperparameters $\theta$~\cite{williams2006gaussian}.
We call $Y$ a Gaussian process if for every finite sample of $Y$,
\begin{equation}
	Y(\bm{X}) \sim \mathcal{N}(\bm{m}(\bm{X}),\bm{K}_{\theta}(\bm{X},\bm{X})).
\end{equation}
$\mathcal{N}$ denotes the multivariate Gaussian distribution, $\bm{m}(\bm{X})$ is the mean vector of size $n$,
and $\bm{K}_{\theta}(\bm{X},\bm{X})$ is the covariance matrix generated by the kernel $k_{\theta}(\bm{x},\bm{x'})$.
In this manuscript we assume that $\bm{K}_{\theta}{(\bm{X},\bm{X})}$ is induced by the Mat\'ern kernel $\phi_{\gamma^2,\rho,\nu,\tau}$, where for points $\bm{x}, \bm{x^\prime}$ where $\| \bm{x} - \bm{x^\prime} \|_2 = d$,
\begin{equation}
	k_{\gamma^2,\rho,\nu,\tau}(\bm{x}, \bm{x^\prime}) 
	= \phi_{\gamma^2,\rho,\nu,\tau}(d) 
	= \gamma^2 \left[ \frac{2^{1-\nu}}{\Gamma(\nu)} \left( \sqrt{2\nu}\frac{d}{\rho} \right)^{\nu} K_{\nu} \left( \sqrt{2\nu}\frac{d}{\rho} \right) + \tau^2\mathbb{I} \left( d = 0 \right) \right].
\end{equation}
Here $\Gamma(\cdot)$ is the gamma function and $K_{\nu}$ is the modified Bessel function of the second kind.

For unobserved points $\bm{Z} \in \mathbb{R}^{m \times d}$, we predict the response distribution with mean $Y(\bm{Z})$ and variance $\textrm{Var}(Y(\bm{Z}))$ where
\begin{align}
	Y(\bm{Z}) \approx \tilde{Y}_\theta(\bm{Z} \vert \bm{X}) 
	&= \bm{K}_{\theta}(\bm{Z},\bm{X})\bm{K}_{\theta}{(\bm{X},\bm{X})}^{-1}Y(\bm{X}), \label{eqn:post_mean} \\
	\textrm{Var}(Y(\bm{Z})) \approx \textrm{Var}(\tilde{Y}_\theta(\bm{Z} \vert \bm{X})) 
	&= \bm{K}_\theta(\bm{Z}, \bm{Z}) - \bm{K}_{\theta}(\bm{Z},\bm{X})\bm{K}_{\theta}{(\bm{X},\bm{X})}^{-1} \bm{K}_{\theta}(\bm{X}, \bm{Z}). \label{eqn:post_var}
\end{align}
Here $\bm{K}_{\theta}(\bm{Z},\bm{X}) = \bm{K}_{\theta}(\bm{X},\bm{Z})^\top$ is the cross covariance of the test points $\bm{Z}$ and training data $\bm{X}$.

\textbf{MuyGPs hyperparameter optimization.}
Conventional GP training consists of maximizing the log-likelihood of the training data $Y(\bm{X})$ given $\theta$, requiring $\mathcal{O}(n^3)$ FLOPs and $\mathcal{O}(n^2)$ storage, which is prohibitively expensive in large-scale applications.
Scalable GP algorithms, e.g.~\cite{vecchia1988estimation,nguyen2009model} seek to address this computational bottleneck (see~\cite{liu2020gaussian} for an extensive review).
MuyGPs is a global approximation algorithm that accelerates hyperparameter optimization by limiting
the kernel matrix to the nearest neighbor structure of the training data (see~\cite{datta2016hierarchical,datta2016nearest}), batching, and replacing expensive
log-likelihood evaluations with leave-one-out cross-validation (LOOCV)~\cite{muyskens2021muygps}. LOOCV withholds
the $i$th training location $\bm{x}_i$ and predicts its response $Y(\bm{x}_i)$ using the other $n-1$ points.
MuyGPs conditions a training feature vector $\mathbf{x}_i$ on its $k$ nearest neighbors, denoted $\bm{X}_{N_i}$, yielding the
prediction
\begin{align}
	\hat{Y}_{\theta}(\bm{x}_i \vert \bm{X}_{N_i}) 
	&= \bm{K}_{\theta}(\bm{x}_i,\bm{X}_{N_i}) {\bm{K}_{\theta}(\bm{X}_{N_i},\bm{X}_{N_i})}^{-1}Y(\bm{X}_{N_i}), \label{eqn:train_mean} \\
	\text{Var}(\hat{Y}_\theta(\bm{x}_i \mid \bm{X}_{N_i}))
	&= \bm{K}_\theta(\bm{x}_i, \bm{x}_i) - \bm{K}_\theta(\bm{x}_i, \bm{X}_{N_i}) \bm{K}_\theta(\bm{X}_{N_i}, \bm{X}_{N_i})^{-1} \bm{K}_\theta(\bm{X}_{N_i}, \bm{x}_i). \label{eqn:train_var}
\end{align}
The MuyGPs training procedure minimizes a loss function $Q(\theta)$ over a randomly sampled batch of training
points $B$ with $b = \vert B \vert$. Training $\theta$ amounts to minimizing $Q(\theta)$ with respect to $\theta$:
\begin{equation} \label{eqn:q}
	\hat{\theta} \in \argmin_{\theta} Q(\theta).
\end{equation}
Using loss functions such as MSE and leave-one-out log-likelihood (LOOL)~\cite{sundararajan1999predictive}, evaluating Equation~\eqref{eqn:q} requires $\mathcal{O}(bk^3)$ FLOPS.
This is much cheaper than the $\mathcal{O}(n^3)$ cost of log-likelihood maximization.
MuyGPs predicts the response distribution for a novel point $\bm{z}$ with neighbors $\bm{X}_{N^*}$,
\begin{align}
	\hat{Y}_{\hat{\theta}}(\bm{z} \vert \bm{X}) 
	&= \bm{K}_{\hat{\theta}}(\bm{z},\bm{X}_{N^*})\bm{K}_{\hat{\theta}}{(\bm{X}_{N^*},\bm{X}_{N^*})}^{-1}Y(\bm{X}_{N^*}), \label{eqn:muygps_mean} \\
	\text{Var}(\hat{Y}_{\hat{\theta}}(\bm{z} \mid \bm{X}))
	& = \bm{K}_{\hat{\theta}}(\bm{z}, \bm{z}) - \bm{K}_{\hat{\theta}}(\bm{z}, \bm{X}_{N^*}) \bm{K}_{\hat{\theta}}(\bm{X}_{N^*}, \bm{X}_{N^*})^{-1} \bm{K}_{\hat{\theta}}(\bm{X}_{N^*}, \bm{z}). \label{eqn:muygps_var}
\end{align}

\textbf{Hyperparameter optimization with LOOL and coverage.}\label{sec:hyperparametertraining}
The success of the MuyGPs method lies in the combination of LOOCV and nearest-neighbor approximations.
Hence, the LOOL is a natural choice of criterion as it allows us to incorporate both of these features while retaining the predictions and variance.
We formulate the LOOL loss function (excluding the constant term) computed using LOOCV and local Kriging on the batched training examples via
\begin{equation} \label{eqn:q_multi_1}
	%Q_1(\theta) = \sum_{i \in B} {-\frac{1}{2}\log(2\pi) -\frac{1}{2}\log(\sigma_i^2) - \frac{(Y(\bm{x}_i) - \hat{Y}_{\theta}(\bm{x}_i \vert \bm{X}_{N_i}))^2}{2\sigma_i^2}} .
	Q(\theta) = \sum_{i \in B} {\log(\sigma_i^2(\theta)) + \frac{(Y(\bm{x}_i) - \mu_i(\theta))^2}{\sigma_i^2(\theta)}},
\end{equation}
where $\mu_i(\theta)$ and $\sigma_i^2(\theta)$ are the posterior mean and variance of the $i$th batch point as defined in Equations \eqref{eqn:train_mean} and \eqref{eqn:train_var}, respectively.

%Unfortunately, the Mat\'ern kernel hyperparameters $\rho$ and $\nu$ used implicitly in constructing $\mu_i(\theta)$ and $\sigma_i^2(\theta)$ are mutually non-identifiable~\cite{stein1999interpolation}.
We augment Equation \eqref{eqn:q_multi_1} with a multi-level coverage penalty.
Let $z_{\alpha}$ be a z-score corresponding to a given confidence level $\alpha$, e.g., $z_{0.95}=1.96$. 
Then, the coverage function $c_{\alpha}(\theta)$ is given by the fraction of ground truth response values for $i \in B$ which lie with a confidence interval of width $z_{\alpha}\sigma_i(\theta)$ around $\mu_i(\theta)$, written as

\begin{equation} \label{eqn:coverage}
	c_{\alpha}(\theta) = \frac{1}{b} \sum_{i \in B} \bm{1}_{\left( \mu_i(\theta)- z_{\alpha} \sigma_i(\theta) ,\ \mu_i(\theta) +  z_{\alpha} \sigma_i(\theta) \right)}(Y(\bm{x}_i) ) .
\end{equation}

We can tune the statistical coverage of the model by constraining Equation~\eqref{eqn:q_multi_1} with Equation~\eqref{eqn:coverage}.
For example, we can tune $\rho$ to ensure that 95 percent of the responses $Y(\bm{x}_i)$ are within 1.96 standard deviations of the posterior mean of the trained GP, similar to conformal prediction algorithms~\cite{vovk2005algorithmic}.

\textbf{LOOL with a coverage penalty.}
We introduce a sequence of $m$ confidence levels $\{\alpha_{j}\}_{j=1}^{m}$.
The coverage at these values will serve as a penalty on the LOOL.
% This yields the problem:
% \begin{equation}
% 	\label{eq:constrained_LOOL}
% 	\begin{aligned}
% 		\min_{\theta} \quad & Q (\theta) ,                                               \\
% 		\textrm{s.t} \quad  & c_{\alpha_{j}} (\theta) = \alpha_{j}, \quad j=1,\hdots, m.
% 	\end{aligned}
% \end{equation}
% We note that in general this optimization is non-convex, and derivatives, even when available, are prohibitively expensive to compute. 
We employ a combination of method of multipliers and Bayesian optimization to accommodate the lack of derivatives.
Method of multipliers formulates the problem by introducing a quadratic penalty on the objective weighted by a parameter $\beta> 0$ \cite{hestenes1969multiplier}.
If we denote the vectorized coverage and confidence level quantities as $C_{\alpha}(\theta) = [c_{\alpha_{j}}(\theta)]_{j=1}^{m}$ and $\alpha=[\alpha_{j}]_{j=1}^{m}$ respectively, then this new problem can be written as:
\begin{equation}
	\label{eq:quadratic_penalty}
	\begin{aligned}
		\min_{\theta} \quad & Q (\theta) +\frac{\beta}{2}\Vert C_{\alpha}(\theta)-\alpha \Vert^{2}_{2} , \\
		\textrm{s.t} \quad  & C_{\alpha} (\theta) = \alpha.
	\end{aligned}
\end{equation}

We formulate the augmented Lagrangian to incorporate the penalty, 
\begin{equation}
	\label{eq:augmented_lagrangian}
	\mathcal{L} (\theta,\lambda;\beta) = Q (\theta) + \langle \lambda , C_{\alpha}(\theta) - \alpha \rangle
	+ \frac{\beta}{2}\Vert C_{\alpha}(\theta)-\alpha \Vert^{2}_{2} ,
\end{equation}
and employ method of multipliers to update the hyperparameters $\theta$ and Lagrange multipliers $\lambda$.
%A common practice is to increase $\beta$ at each iteration to accelerate convergence. For this reason, we choose a geometrically increasing
%sequence $\beta_{n} = r^{n}\beta_{0}$ at iteration $n$ where $\beta_{0}>0$ and $r>1$.

\begin{algorithm}[t]
	\caption{Scalable GP Hyperparameter Optimization via Method of Multipliers}\label{alg:muygps}
	\begin{algorithmic}[1]
		\Procedure{MM$_{train}$}{$k$, $b$, $\bm{X}$, $Y(\bm{X})$, $\theta$}
		\State $b \leftarrow$ batch size; $k \leftarrow$ number of nearest neighbors
		\State $\bm{X}, Y(\bm{X}) \leftarrow$ train features and responses
		\State $\theta_0 \leftarrow$ hyperparameters initial guess
		\State $B \leftarrow$ uniform sample of size b from $\lbrace 1, \dots , n \rbrace$
		\State $\bm{X}_{N_i} \leftarrow$ query $k$ nearest neighbors for all $i \in B$

		\For {$n=1,\dots,N$}
		\State $\theta_{n} \in \argmin_{\theta} \mathcal{L}(\theta,\lambda_{n-1}; \beta_{n-1})$ \quad (Bayesian Optimization)
		\State $\lambda_{n} = \lambda_{n-1} + \beta_{n-1}\left( C(\theta_{n}) - \alpha \right)$
		\State $\beta_{n} = r\beta_{n-1}$
		\EndFor
		\State \textbf{return} $\theta_{N}$ for use in prediction
		\EndProcedure
	\end{algorithmic}
\end{algorithm}

\textbf{Synthetic data experiment.}
We apply our method to data generated from a univariate Gaussian process using points taken from the unit interval $[0,1]$.
We vary the Mat\'ern kernel hyperparameters $\nu$ and $\rho$ to form four different test cases with $(\nu,\rho) = (0.135,0.95), (0.425,0.625), (0.635,0.475) \text{, and } (0.965,0.125)$, respectively.
We report statistical coverage and hyperparameter estimates for three loss functions: mean-squared error (MSE), LOOL (Equation~\eqref{eqn:q_multi_1}) and the augmented Lagrangian (Equation~\eqref{eq:augmented_lagrangian}) for the method-of-multipliers (MM) implementation.
We visualize all results using violin plots~\cite{hintze1998violin}.

\begin{figure}[h!]
	\centering
	\figuretitle{95th Percentile Statistical Coverage Values Across All Datasets}
	\includegraphics[width=0.5\linewidth]{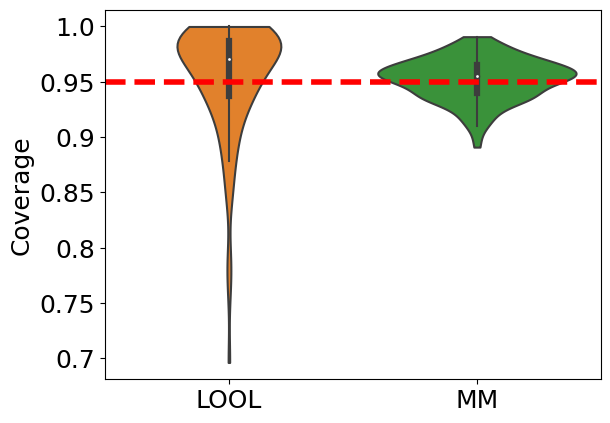}
	\title{BOB}
	\caption{Violin plot of 95th percentile statistical coverage for LOOL and MM. The red dashed line indicates the target coverage value of 95 percent.}
	\label{fig:synthetic_coverage}
\end{figure}

\begin{figure}[h!]
	\centering
	\figuretitle{Estimated Values of $\nu$ (Top) and $\rho$ (Bottom) Across Four Synthetic Datasets}
	\includegraphics[width=0.24\linewidth]{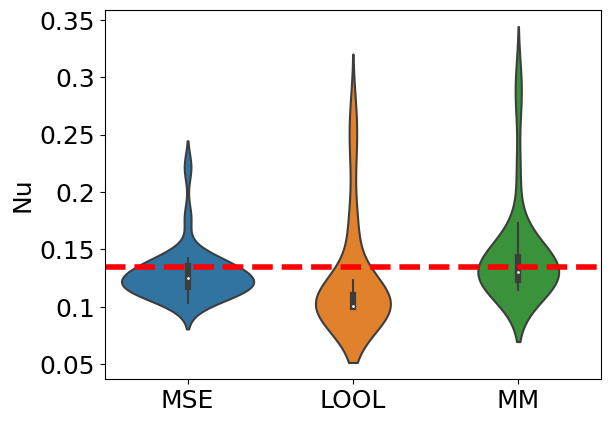}
	\includegraphics[width=0.24\linewidth]{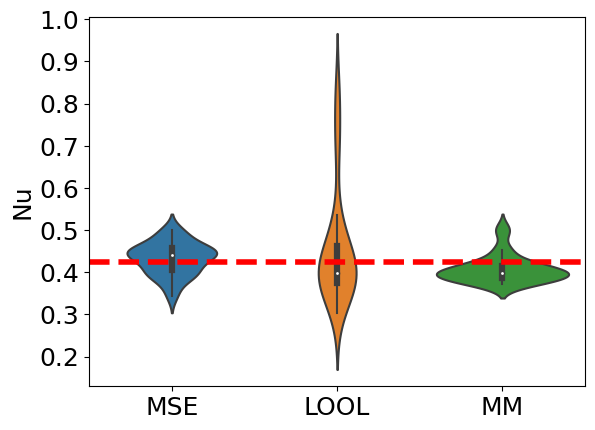}
	\includegraphics[width=0.24\linewidth]{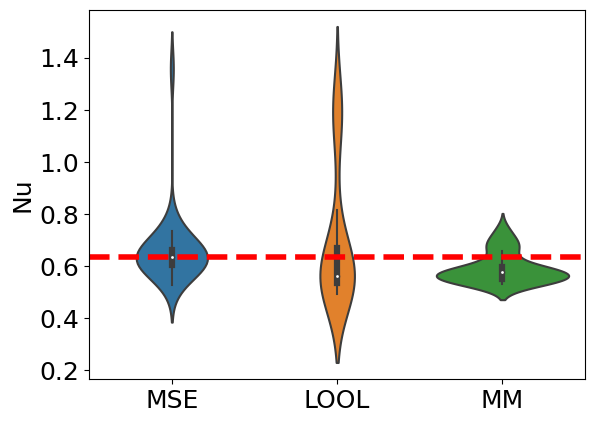}
	\includegraphics[width=0.24\linewidth]{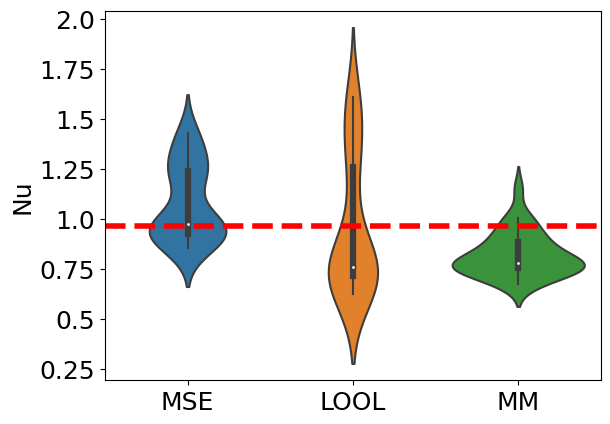}
	\includegraphics[width=0.24\linewidth]{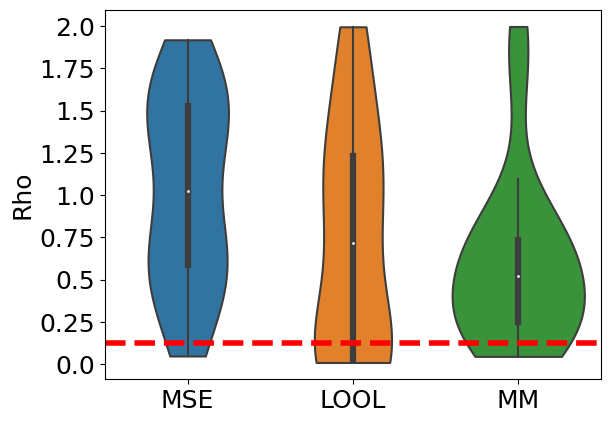}
	\includegraphics[width=0.24\linewidth]{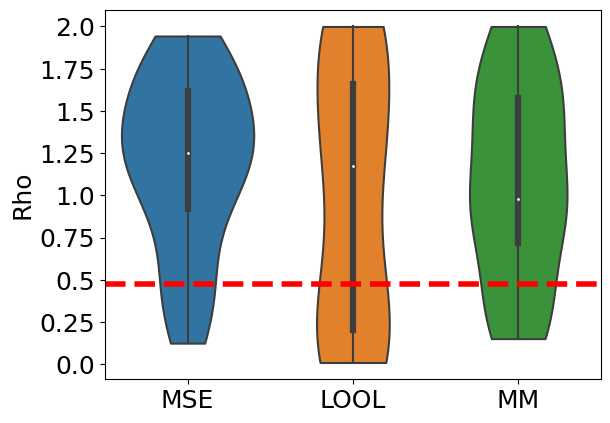}
	\includegraphics[width=0.24\linewidth]{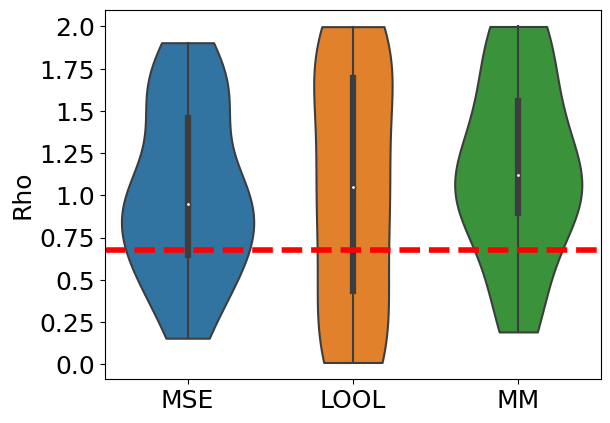}
	\includegraphics[width=0.24\linewidth]{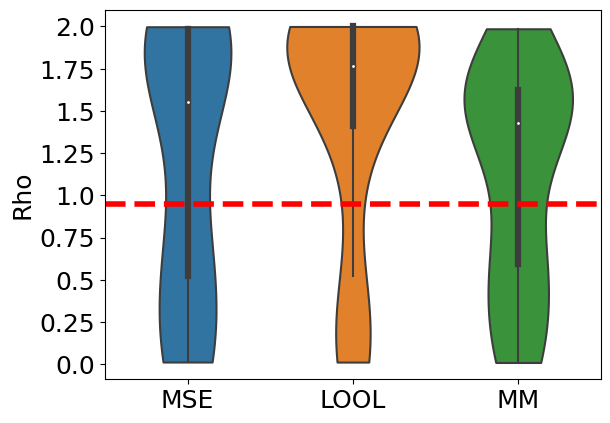}
	\caption{MSE is shown in the left blue violins, LOOL in the center orange, and MM in the right green. The red dashed lines indicate the true hyperparameter value.}
	\label{fig:synthetic_nu_rho}
\end{figure}

In Figure~\ref{fig:synthetic_coverage}, we provide the distribution of coverage values across all trials and datasets.
% The wide range of coverage values in the left panel show that MSE is unreliable for capturing the uncertainty in the response.
LOOL and MM perform quite well in covering the response to the correct extent (95 percent in this case).
Critically, MM achieves coverage closer to the target of 95 percent with much lower variance than LOOL, whereas LOOL tends to overestimate the desired coverage.
This indicates that the coverage-based regularization approach is indeed improving the UQ of the GP predictor.
In the top row of Figure~\ref{fig:synthetic_nu_rho} we observe that the MSE, LOOL, and MM approaches generate close approximations to the smoothness parameter $\nu$.
Interestingly, the MSE and LOOL outperform MM in this case.
This is likely due to the biasing imposed by incorporating the coverage penalty, but does not significantly negatively impact predictive performance.
In the bottom row of Figure~\ref{fig:synthetic_nu_rho} we observe that all three methods give poor estimates of the length scale parameter $\rho$, reflecting the mutual non-identifiability of $\nu$ and $\rho$ in the Mat\'ern kernel~\cite{stein1999interpolation}.

\begin{figure}[h!]
	\centering
	\figuretitle{Performance Metrics for LOOL and MM on the Surface Temperature Dataset}
	\includegraphics[width=0.19\linewidth]{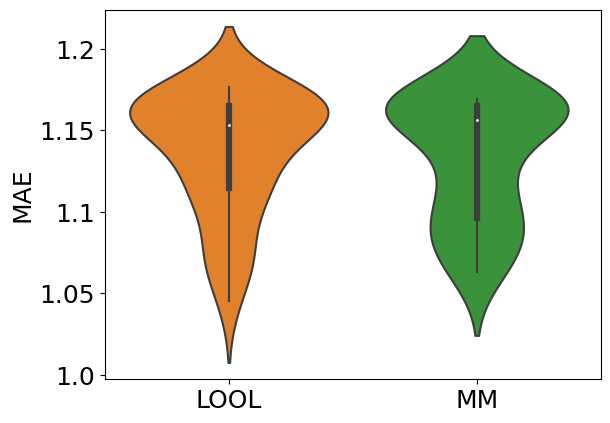}
	\includegraphics[width=0.19\linewidth]{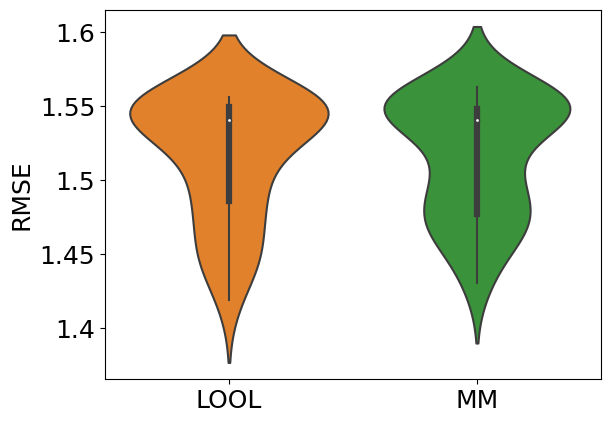}
	\includegraphics[width=0.19\linewidth]{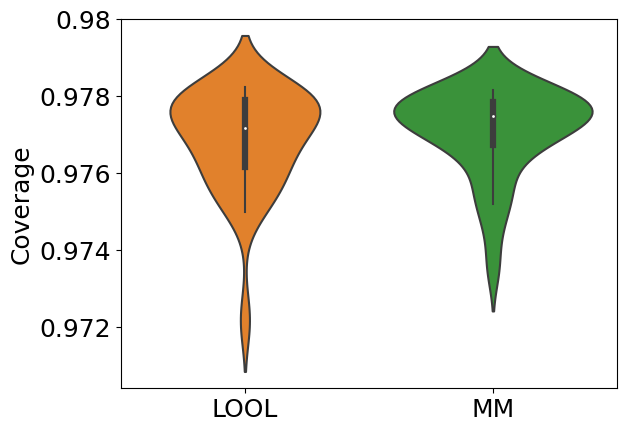}
	\includegraphics[width=0.19\linewidth]{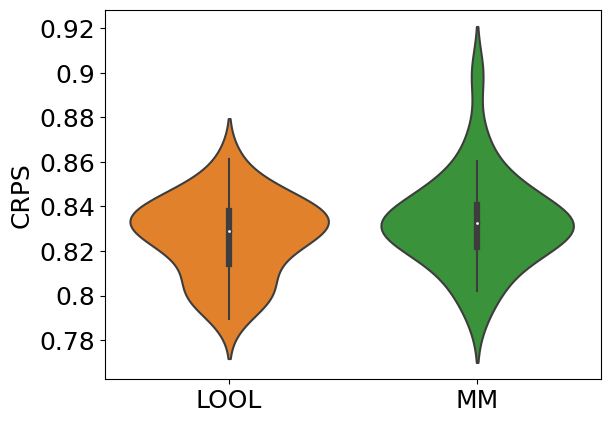}
	\includegraphics[width=0.19\linewidth]{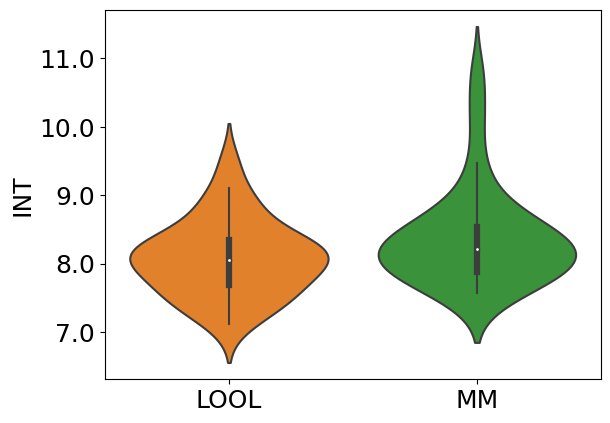}
	\caption{From left to right: the mean absolute error (MAE), root MSE, 95th percentile statistical coverage, continuous rank probability score (CRPS)~\cite{gneiting2007strictly}, and interval score (INT)~\cite{gneiting2007strictly} for surface temperature dataset from the GP competition paper~\cite{heaton2019case}.}
	\label{fig:heaton}
\end{figure}

\textbf{Ground surface temperature data experiment.} We now study a dataset comprising land surface temperatures measured by a Terra instrument from the MODIS satellite on August 4, 2016 on a $500 \times 300$ grid between longitudes -95.91153 and -91.28381 and latitudes 34.29519 to 37.06811, with 105,569 training observations and 42740 testing observations left over after removing unmeasured points (see~\cite{heaton2019case} for more detail).
In this case we only compare LOOL and MM, as these two methods achieved significantly better results than the MSE loss function.
Figure~\ref{fig:heaton} provides the mean absolute error (MAE), root MSE, 95th percentile statistical coverage (COV), continuous rank probability score (CRPS)~\cite{gneiting2007strictly}, and interval score (INT)~\cite{gneiting2007strictly}.
MM and LOOL achieve similar performance metrics on this test problem.
As the optimal value of $\nu$ in this case is close to 1, the coverage regularization is less effective than it is in the small $\nu$ regime (see Supplementary Material).
However, both MM and LOOL impressively outperform all methods in the competition paper~\cite{heaton2019case} and the original MuyGPs algorithm in~\cite{muyskens2021muygps}.
This test case demonstrates the scalability of the coverage regularization technique and its applicability to large-scale real-world datasets.

\textbf{Conclusions, limitations, \& future work.}
We have presented a novel hyperparameter estimation algorithm for improved UQ in Gaussian process regression.
Our approach demonstrates meaningful improvement in statistical coverage and other UQ-centric performance metrics over a leave-one-out likelihood maximization approach.
As demonstrated in the second test case, the algorithm is highly scalable; it trains GP hyperparameters on problems with > 100,000 data points on a standard laptop.
The experiments presented in this paper are limited in extent; a more thorough comparison of our approach to state-of-the-art GP hyperparameter estimation algorithms, as well as runtime analysis, is a necessary next step.
Future extensions of this work could also include exploration of other loss functions and constraints based on methods from Conformal Prediction~\cite{vovk2005algorithmic,shafer2008tutorial}.

\section*{Acknowledgments}
This work was performed under the auspices of the U.S. Department of Energy by Lawrence Livermore National Laboratory under Contract DE-AC52-07NA27344 with IM release number LLNL-CONF-839970. 
Funding for this work was provided by LLNL Laboratory Directed Research and Development grant 22ERD028.
\bibliographystyle{plain}
% \bibliography{paper}

\begin{center}
	{\bf \Large Supplementary Material}
\end{center}

\appendix
\section{Experimental details}
We provide detailed descriptions of the datasets used in the numerical experiments.
We also provide information on the train/test splits and training hyperparameters used in the test cases.
We run 30 trials for all dataset-algorithm combinations.

\subsection{Synthetic data}
In the synthetic datasets, we use a GP with pre-fixed hyperparameters and a nugget parameter of $\tau^2 = 10^{-10}$ to generate 10000 data points.
We use a 50/50 train-test split, yielding 5000 training samples and 5000 test samples.
During training, we fix $\gamma^2$ at 1.
In the hyperparameter optimization, we use 50 nearest neighbors, a batch size of 1024, and confidence levels of $0.90,0.925,0.95,0.975,\text{ and } 0.99$ to construct the regularization term.
In the LOOL optimization, the Bayesian optimization procedure uses 5 initial points, 30 iterations, an expected improvement acquisition function, and an exploration parameter $\kappa = 7$.
In the dual-ascent augmented Lagrangian approach, the Bayesian optimization procedure uses 3 initial points, 10 iterations, an expected improvement acquisition function, and an exploration parameter $\kappa = 7$.

We provide detailed results for the synthetic data experiments in 4 separate tables below.
We compute the mean and standard deviation of each value across 30 trials.
\begin{center}
	\begin{table}[h]
		\begin{tabular}{ |p{2cm}p{3cm}p{3cm}p{3cm}|  }
			\hline
			\multicolumn{4}{|c|}{Performance metrics for $\nu = 0.135$, $\rho = 0.95$}                  \\
			\hline
			Loss Function    & MSE                    & LOOL                   & MM                     \\
			Estimated $\nu$  & $0.129 \pm  2.27e$-$2$ & $0.124 \pm  4.80e$-$2$ & $0.145 \pm  4.40e$-$2$ \\
			Estimated $\rho$ & $1.18 \pm 0.795 $      & $1.50 \pm  0.705 $     & $1.11 \pm  0.649 $     \\
			MAE              & $0.237 \pm 4.01e$-$4$  & $0.238 \pm  8.54e$-$4$ & $0.238 \pm  1.31e$-$3$ \\
			RMSE             & $0.299 \pm 4.99e$-$4$  & $0.300 \pm  1.03e$-$2$ & $0.299 \pm  1.57e$-$3$ \\
			COV              & $0.943 \pm 8.35e$-$2$  & $0.970 \pm  3.28e$-$2$ & $0.942 \pm  1.82e$-$2$ \\
			CRPS             & $0.174 \pm 8.56e$-$3$  & $0.171 \pm  8.32e$-$4$ & $0.169 \pm  1.14e$-$3$ \\
			INT              & $1.65 \pm 0.464 $      & $1.50 \pm  3.76e$-$2$  & $1.41 \pm  3.17e$-$2$  \\
			\hline
		\end{tabular}
	\end{table}
\end{center}

\begin{center}
	\begin{tabular}{ |p{2cm}p{3cm}p{3cm}p{3cm}|  }

		\hline
		\multicolumn{4}{|c|}{Performance metrics for $\nu = 0.425$, $\rho = 0.675$}                             \\
		\hline
		Loss Function    & MSE                        & LOOL                       & MM                         \\
		Estimated $\nu$  & $0.433 \pm  3.83e$-$2$     & $0.453 \pm  0.133 $        & $0.408 \pm  3.24e$-$2$     \\
		Estimated $\rho$ & $1.02\pm 0.493 $           & $1.02 \pm 0.700 $          & $1.16 \pm 0.526 $          \\
		MAE              & $ 1.88e$-$2 \pm 1.46e$-$5$ & $ 1.89e$-$2 \pm 1.21e$-$4$ & $ 1.88e$-$2 \pm 1.24e$-$5$ \\
		RMSE             & $ 2.36e$-$2 \pm 1.71e$-$5$ & $ 2.37e$-$2 \pm 1.37e$-$4$ & $ 2.36e$-$2 \pm 1.13e$-$5$ \\
		COV              & $0.861 \pm 0.126 $         & $0.950 \pm 4.92e$-$2$      & $0.949 \pm 1.10e$-$2$      \\
		CRPS             & $ 1.38e$-$2 \pm 5.75e$-$4$ & $ 1.36e$-$2 \pm 4.74e$-$4$ & $ 1.33e$-$2 \pm 1.05e$-$5$ \\
		INT              & $0.147 \pm 4.72e$-$2$      & $0.121 \pm 1.73e$-$2$      & $0.111 \pm 4.86e$-$4$      \\
		\hline
	\end{tabular}
\end{center}

\begin{center}
	\begin{tabular}{ |p{2cm} p{3cm} p{3cm} p{3cm}|  }

		\hline
		\multicolumn{4}{|c|}{Performance metrics for $\nu = 0.635$, $\rho = 0.475$}                             \\
		\hline
		Loss Function    & MSE                        & LOOL                       & MM                         \\
		Estimated $\nu$  & $0.661 \pm  0.141 $        & $0.696 \pm  0.261 $        & $0.590 \pm  5.82e$-$2$     \\
		Estimated $\rho$ & $1.19\pm 0.510 $           & $0.985 \pm 0.765 $         & $1.09 \pm 0.583 $          \\
		MAE              & $ 3.83e$-$3 \pm 1.57e$-$5$ & $ 3.84e$-$3 \pm 2.64e$-$5$ & $ 3.82e$-$3 \pm 2.07e$-$6$ \\
		RMSE             & $ 4.79e$-$3 \pm 1.71e$-$5$ & $ 4.81e$-$3 \pm 2.67e$-$5$ & $ 4.79e$-$3 \pm 5.61e$-$6$ \\
		COV              & $0.701 \pm 0.251 $         & $0.947\pm 5.65e$-$2$       & $0.953 \pm 1.45e$-$2$      \\
		CRPS             & $ 2.99e$-$3 \pm 3.57e$-$4$ & $ 2.78e$-$3 \pm 1.02e$-$4$ & $ 2.71e$-$3 \pm 4.98e$-$6$ \\
		INT              & $4.73e$-$2 \pm 3.01e$-$2$  & $2.58e$-$2 \pm 4.19e$-$3$  & $2.28e$-$2 \pm 1.87e$-$4$  \\
		\hline
	\end{tabular}
\end{center}

\begin{center}
	\begin{tabular}{ |p{2cm} p{3cm} p{3cm} p{3cm}|  }

		\hline
		\multicolumn{4}{|c|}{Performance metrics for $\nu = 0.965$, $\rho = 0.125$}                          \\
		\hline
		Loss Function    & MSE                       & LOOL                      & MM                        \\
		Estimated $\nu$  & $1.08 \pm  0.190 $        & $0.951 \pm  0.342 $       & $0.820 \pm 0.110$         \\
		Estimated $\rho$ & $1.02\pm 0.561 $          & $0.742 \pm 0.676 $        & $ 0.603 \pm 0.501$        \\
		MAE              & $ 9.02e$-$4 \pm 8e$-$6$   & $ 9.04e$-$4 \pm 4e$-$6$   & $ 9.00e$-$4 \pm 3e$-$6$   \\
		RMSE             & $ 1.13e$-$3 \pm 9e$-$6$   & $ 1.13e$-$3 \pm 5e$-$6$   & $ 1.12e$-$4 \pm 3e$-$6$   \\
		COV              & $0.271 \pm 0.326 $        & $0.950\pm 6.37e$-$2$      & $0.965 \pm 2.32e$-$2$     \\
		CRPS             & $ 8.28e$-$4 \pm 9.5e$-$5$ & $ 6.53e$-$4 \pm 2.9e$-$5$ & $ 6.38e$-$4 \pm 4e$-$6$   \\
		INT              & $2.58e$-$2 \pm 1.05e$-$2$ & $6.01e$-$3 \pm 1.13e$-$3$ & $5.37e$-$3 \pm 1.63e$-$4$ \\
		\hline
	\end{tabular}
\end{center}

\begin{center}
	\begin{tabular}{ |p{2cm} p{3cm} p{3cm} p{3cm}|  }

		\hline
		\multicolumn{4}{|c|}{Coverage across all four datasets}                            \\
		\hline
		Loss Function & MSE                & LOOL                  & MM                    \\
		COV           & $0.694 \pm 0.338 $ & $0.954 \pm 5.20e$-$2$ & $0.953 \pm 1.90e$-$2$ \\
		\hline
	\end{tabular}
\end{center}

\subsection{Heaton et al. dataset}
For the Heaton et al. dataset, we use a GP with a nugget parameter of $\tau^2 = 10^{-3}$.
In the hyperparameter optimization, we use 50 nearest neighbors, a batch size of 1024, and confidence levels of $0.90,0.925,0.95,0.975,\text{ and } 0.99$ to construct the regularization term.
In the LOOL optimization, the Bayesian optimization procedure uses 5 initial points, 30 iterations, an expected improvement acquisition function, and an exploration parameter $\kappa = 7$.
In the dual-ascent augmented Lagrangian approach, the Bayesian optimization procedure uses 3 initial points, 10 iterations, an expected improvement acquisition function, and an exploration parameter $\kappa = 7$.
The dataset was constructed from the repository at \hyperlink{heatonlink}{https://github.com/finnlindgren/heatoncomparison}.

We provide detailed results for the Heaton et al. dataset experiments.
We compute the mean and standard deviation of each value across 30 trials.

\begin{center}
	\begin{tabular}{ |p{2cm} p{3cm} p{3cm}|  }

		\hline
		\multicolumn{3}{|c|}{Performance metrics for surface temperature dataset}  \\
		\hline
		Loss Function    & LOOL                      & MM                          \\
		Estimated $\nu$  & $0.846 \pm  0.105$        & $0.889 \pm  7.57e$-$2$      \\
		Estimated $\rho$ & $1.14e$-$2 \pm 1.98e$-$3$ & $ 1.17e$-$2  \pm 2.31e$-$3$ \\
		MAE              & $ 1.14 \pm 3.69e$-$2$     & $ 1.13 \pm 3.81e$-$2$       \\
		RMSE             & $ 1.52 \pm 4.16e$-$2$     & $ 1.52 \pm 4.03e$-$2$       \\
		COV              & $0.977 \pm 1.32e$-$3$     & $0.977 \pm 1.13e$-$3$       \\
		CRPS             & $ 0.826 \pm 1.77e$-$2$    & $ 0.833 \pm 2.11e$-$2$      \\
		INT              & $ 8.08 \pm 0.553$         & $8.37 \pm 0.711$            \\
		\hline
	\end{tabular}
\end{center}

\end{document}